\documentclass[letterpaper]{article} 
\usepackage{aaai2026}  
\usepackage{times}  
\usepackage{helvet}  
\usepackage{courier}  
\usepackage[hyphens]{url}  
\usepackage{graphicx} 
\usepackage{natbib}  
\usepackage{caption} 
\usepackage{algorithm}
\usepackage{algpseudocode}
\usepackage[utf8]{inputenc} 
\usepackage[T1]{fontenc}    
\usepackage{hyperref}       
\usepackage{url}            
\usepackage{booktabs}       
\usepackage{amsfonts}       
\usepackage{nicefrac}       
\usepackage{microtype}      
\usepackage{xcolor}         
\usepackage[utf8]{inputenc} 
\usepackage[T1]{fontenc}    
\usepackage{url}            
\usepackage{booktabs}       
\usepackage{amsfonts}       
\usepackage{nicefrac}       
\usepackage{microtype}      
\usepackage{colortbl}
\usepackage{lipsum} 
\usepackage{tcolorbox}
\tcbuselibrary{listings, skins, breakable}
\usepackage{amssymb}
\usepackage{booktabs}  
\usepackage{multirow}  
\usepackage{tabularx}  
\usepackage{graphicx} 
\usepackage{caption}
\usepackage{amsmath,amssymb}  
\usepackage{adjustbox}
\definecolor{myblue}{RGB}{220,230,242}  
\definecolor{mygreen}{RGB}{226,240,217} 

\algrenewcommand\alglinenumber[1]{\footnotesize #1:}
\usepackage{newfloat}
\usepackage{listings}
\DeclareCaptionStyle{ruled}{labelfont=normalfont,labelsep=colon,strut=off} 
\floatstyle{ruled}
\newfloat{listing}{tb}{lst}{}
\floatname{listing}{Listing}
\nocopyright
\title{TIDE : Temporal-Aware Sparse Autoencoders for Interpretable Diffusion Transformers in Image Generation}
\author{
    Victor Shea-Jay Huang\equalcontrib\textsuperscript{\rm 1,2},
    Le Zhuo\equalcontrib\textsuperscript{\rm 1,2},
    Yi Xin\textsuperscript{\rm 2,3},
    Zhaokai Wang\textsuperscript{\rm 2,4},
    Fu-Yun Wang\textsuperscript{\rm 1},
    Yuchi Wang\textsuperscript{\rm 1},
    Renrui Zhang\textsuperscript{\rm 1},
    Peng Gao\textsuperscript{\rm 2},
    Hongsheng Li\thanks{Corresponding author}\textsuperscript{\rm 1}
}
\affiliations{
    \textsuperscript{\rm 1}CUHK, MMLab 
    \textsuperscript{\rm 2}Shanghai AI Laboratory 
    \textsuperscript{\rm 3}NJU 
    \textsuperscript{\rm 4}SJTU \\
    jeix782@gmail.com, hsli@ee.cuhk.edu.hk
}

\usepackage{bibentry}

\begin{document}

\maketitle

\begin{abstract}

Diffusion Transformers (DiTs) are a powerful yet underexplored class of generative models compared to U-Net-based diffusion architectures. We propose TIDE—Temporal-aware sparse autoencoders for Interpretable Diffusion transformErs—a framework designed to extract sparse, interpretable activation features across timesteps in DiTs. TIDE effectively captures temporally-varying representations and reveals that DiTs naturally learn hierarchical semantics (e.g., 3D structure, object class, and fine-grained concepts) during large-scale pretraining. Experiments show that TIDE enhances interpretability and controllability while maintaining reasonable generation quality, enabling applications such as safe image editing and style transfer.
\end{abstract}

\begin{figure*}[htbp]  
\setlength{\belowcaptionskip}{-15pt}  
  \centering
  \includegraphics[width=0.9\textwidth]{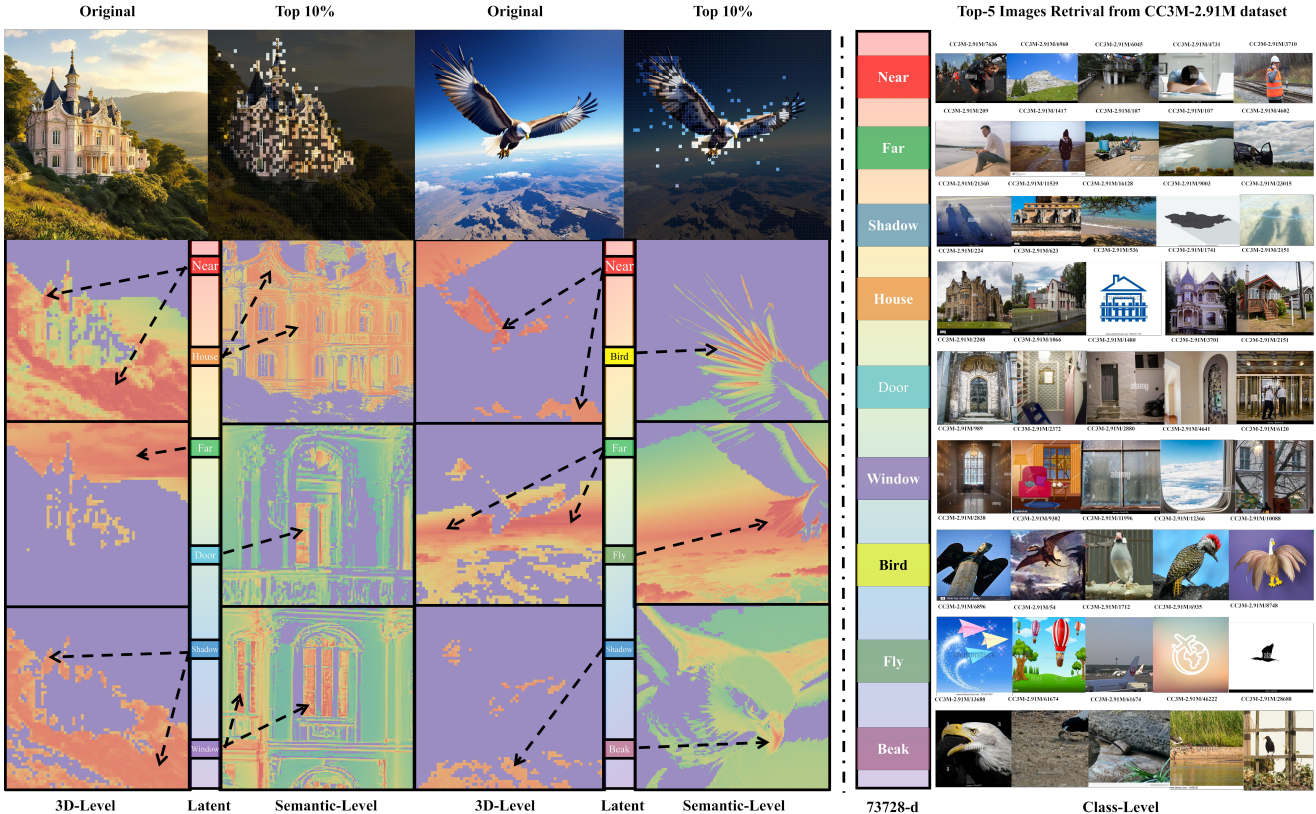}   
  \caption{
TIDE effectively encodes interpretable features across different levels (class, semantic, and 3D) in pre-trained diffusion transformer~\cite{chen2023pixart}. This validates that the diffusion model inherently captures and organizes these multi-level concepts by large-scale 
generative pre-training, enabling it to perform various downstream diffusion tasks effectively(see Sec.Diffusion Really Learned Features for details).
  }
  \label{fig:mask}
\end{figure*}

\vspace{-0.5cm}
\section{Introduction}
\label{sec:intro }

Recent advancements in diffusion models have revolutionized text-to-image generation, with Diffusion Transformers (DiTs) \citep{dit} emerging as a scalable alternative to traditional U-Net backbones \citep{ronneberger2015u} by leveraging transformer-based architectures. Improvements in sampling efficiency \citep{yi2024towards} and model scaling \citep{chen2023pixart,esser2024scaling,zhuo2024lumina,gaolumina} 
have further elevated their performance to a new level. However, a critical question remains: how do these models function internally? While interpretability techniques have been extensively developed for U-Net frameworks \citep{kwon2023diffusion, toker2024diffusion}, similar efforts for DiTs are absent. This is primarily due to the fact that most existing methods are tailored to convolutional architectures and depend on spatially structured feature maps, which are not directly available in the token-based representation of DiTs. More broadly, interpretability research on diffusion models lags significantly behind that of GANs~\cite{goodfellow2014generative,bau2018gan,ling2021editgan}, despite the growing prominence of diffusion models in generative modeling. This gap largely stems from diffusion models' reliance on latent space transformations across various denoising levels, which complicates efforts to analyze their internal mechanisms. In particular, little attention has been paid to understanding latent activations within these models, leaving a significant void in our ability to interpret their processes.

Feature interpretability, which aims to uncover how neural networks encode functionality through internal representations \citep{elhage2021framework}, has seen notable success in language models, especially through the use of sparse autoencoders (SAEs)~\cite{ng2011sparse}. These tools disentangle polysemantic neurons into sparse, human-interpretable features \citep{bricken2023monosemanticity, marks2024sparse}. However, existing SAE methods face significant limitations when applied to diffusion models, particularly at the level of activation layers. Diffusion models present unique challenges due to their temporal dynamics: unlike language models, which process static tokens in an auto-regressive manner, DiTs iteratively refine latent representations across timesteps, integrating multimodal inputs from text encoders. This iterative nature introduces difficulties such as long token sequences, high token dimensionality, and time-varying activation patterns, especially for high-resolution images. 
Similarly, although SAEs have proven effective in vision-language models \citep{Daujotas2024b} and convolutional networks \citep{gorton2024missing}, they encounter distinct challenges in DiTs, including:
\begin{enumerate}
    \item Capturing time-varying activation patterns across diffusion steps;
    \item Balancing sparsity and reconstruction fidelity during iterative denoising;
    \item Aligning sparse features with both visual and text-conditioned semantics.
\end{enumerate}

These challenges underscore the need for new interpretability methods tailored to the temporal and multimodal complexities of DiTs with SAE.

To address these issues, we present \textbf{T}emporal-aware SAEs for \textbf{I}nterpretable \textbf{D}iffusion transform\textbf{E}rs (TIDE), a framework designed to capture temporal variations in activation patterns and extract sparse, interpretable features throughout the denoising process. To ensure high-fidelity feature learning under sparsity constraints, we introduce specialized training strategies, including progressive sparsity scheduling and random sampling augmentation. Additionally, we explore TIDE's scaling laws and validate its performance using diffusion loss, demonstrating that TIDE achieves advanced results compared to other SAE-based methods. Moreover, TIDE effectively learns multi-level features (e.g., 3D, semantics, and class), as shown in Fig.~\ref{fig:mask}, highlighting how diffusion models inherently organize such concepts through large-scale generative pre-training. This finding illustrates their ability to be adapted to various downstream tasks, such as depth estimation~\citep{gui2024depthfm,ke2024repurposing}, semantic segmentation~\citep{tian2024diffuse}, and image classification~\citep{li2023your,mukhopadhyay2023diffusion}. Additionally, we propose a temporal analysis that isolates and interprets the roles of individual timesteps in the generation process, offering deeper insights into the latent representations of text-conditioned image synthesis. By tailoring SAEs to the dynamics of DiTs, TIDE advances model interpretability, enhances understanding of generative mechanisms, and lays the foundation for more trustworthy and controllable generative systems.

This paper presents the first systematic study of SAEs in DiT-based diffusion models. Our contributions are fourfold:
\begin{itemize}
\item \textbf{TIDE Framework}: A paradigm that adapts SAEs to DiT via temporal-aware architecture and progressive sparsity scheduling, coupled with training strategies to extract interpretable features from DiT's latent space while preserving generation quality and balancing sparsity with denoising efficacy.
\item \textbf{Scaling Law and Evaluation}: We analyze the scaling law of TIDE, highlighting its ease of training, and compare its downstream loss with other SAEs to demonstrate the superior performance of TIDE.
\item \textbf{Interpretation}: Using TIDE, we demonstrated that diffusion models learn hierarchical features through generative pre-training, enabling efficient transfer to target domains via fine-tuning. Furthermore, through a decoupled analysis of sampling timesteps, we verify and extend existing understanding about DiT's Coarse-to-Fine generation.
\item \textbf{Potential Applications}: Using PixArt-$\alpha$~\cite{chen2023pixart} and other diffusion models, we demonstrate the practical utility of TIDE-learned sparse features in tasks like safe image editing and style transfer.
\end{itemize}


\begin{figure}[htbp]  
\setlength{\belowcaptionskip}{-15pt}  
  \centering
  \includegraphics[width=\linewidth]{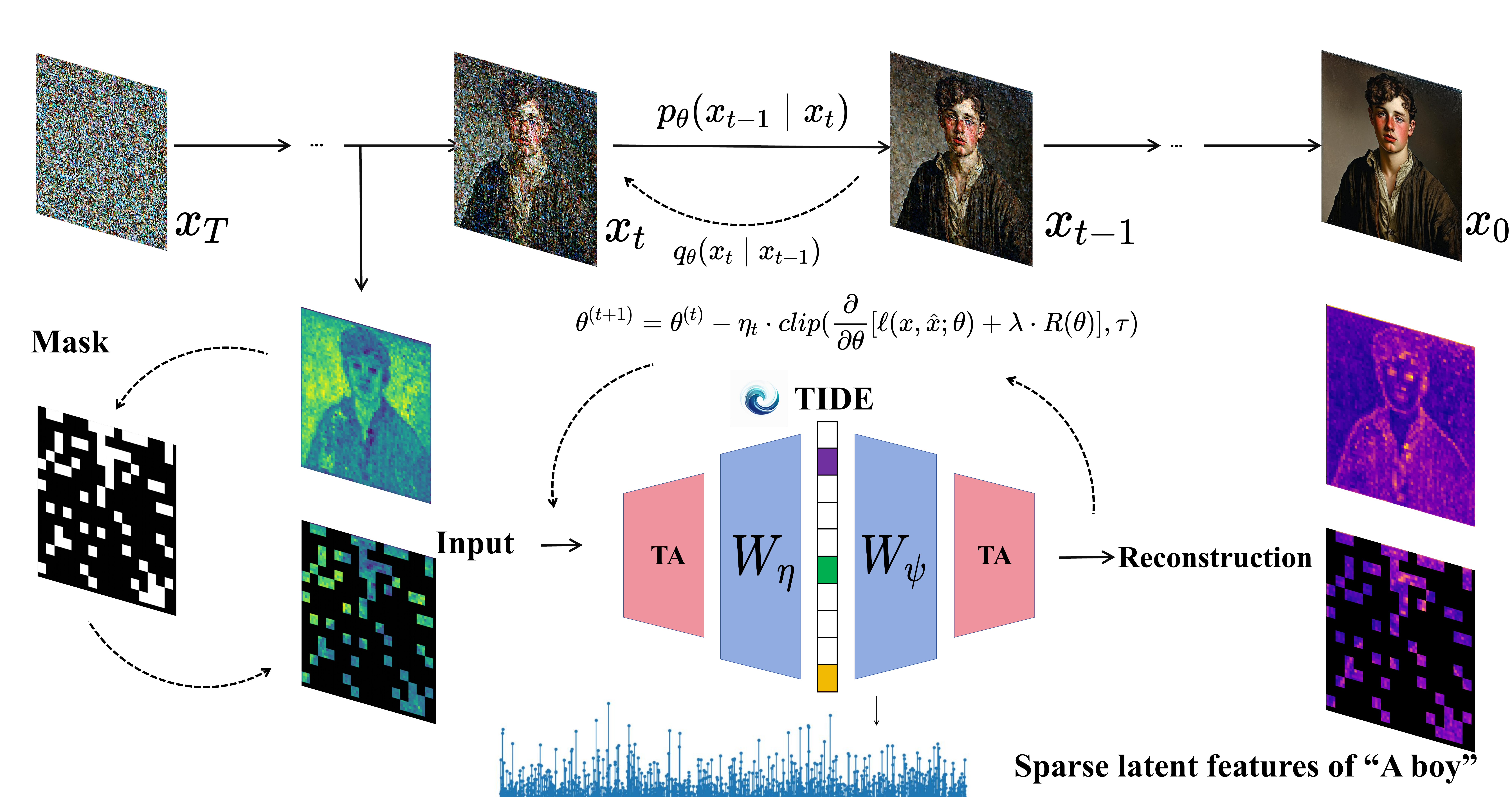}   
  \caption{Overview of our training process: For each Dit, we train its TIDE  (Embed SAE within the TA (temporal-aware) architecture) individually, extracting activation layers from different timesteps each time. The training of the SAE is conducted using both with and without random sampling. For specific details on the loss design, refer to supporting material.}
  \label{fig:train_progress}
\end{figure}

\section{Related work}
\label{sec:Rela}
\subsection{Sparse Autoencoders for Interpretability}
Mechanistic interpretability~\citep{elhage2021framework} aims to understand how neural networks produce outputs by identifying features encoded in neurons~\citep{olah2020zoom}. Sparse autoencoders (SAEs)\citep{Olshausen1997SparseCW} address the polysemantic nature of neurons caused by superposition, where networks encode more features than dimensions\citep{elhage2022toy, sharkey2022superposition}, by enforcing sparsity, activating only a small subset of neurons per input. SAEs have proven effective in uncovering human-interpretable features~\citep{huben2024sparse, bricken2023monosemanticity}, decomposing activations in large language models (LLMs)\citep{templeton2024scaling, yun2021transformer}, and extracting interpretable concepts in vision-language models such as CLIP\citep{Daujotas2024a}. They have also facilitated image editing~\citep{Daujotas2024b}, enabled concept naming~\citep{rao2024discover}, and disabled specific generative features~\citep{bricken2023monosemanticity}, demonstrating their potential to mitigate superposition and enhance interpretability across diverse model architectures.

\subsection{Interpretability of Diffusion Models}

Diffusion models achieve text-to-image generation via iterative denoising, evolving from U-Net architectures~\citep{rombach2022high, podell2023sdxl} to scalable designs like Diffusion Transformer (DiT)~\citep{dit}, with improved sampling efficiency~\citep{song2019generative, song2020denoising, lu2022dpm}. Interpretability efforts focus on bottleneck layers~\citep{kwon2023diffusion, park2023understanding} and cross-attention~\citep{tang2022daam}, enabling manipulation of attributes~\citep{basu2023localizing} and text encoders~\citep{toker2024diffusion}. SAEs, applied to CLIP~\citep{radford2021learning, fry2024towards} and diffusion models~\citep{ijishakin2024h, surkov2024unpacking, kim2024revelio,shabalin2025interpreting}, help identify features, reduce hallucinations~\citep{jiang2024interpreting}, and analyze activations, linking features to visual semantics.
For more details on related work, please refer to the supplementary material.

\section{Preliminaries}
\label{Sec:Preliminaries}

\subsection{Sparse Autoencoder}
\label{subsec:baseine}
To encode the extracted activations into a sparse latent representation, we employ a $k$-sparse autoencoder. The encoder and decoder are defined as follows:

\begin{equation}
\begin{aligned}
    l &= \text{ReLU}(W_\text{enc} a + b_\text{enc}), \\
    \hat{a} &= W_\text{dec} l,
\end{aligned}
\end{equation}
where the encoder weights \( W_\text{enc} \in \mathbb{R}^{n \times f} \), encoder bias \( b_\text{enc} \in \mathbb{R}^{n} \), and decoder weights \( W_\text{dec} \in \mathbb{R}^{f \times n} \) define the parameters of the autoencoder. Here, \( a \in \mathbb{R}^{f} \) is the input vector, \( l \in \mathbb{R}^{n} \) is the latent feature representation, and \( \hat{a} \in \mathbb{R}^{f} \) is the reconstructed vector.

\subsection{TopK Activation}
We employ the $k$-sparse autoencoder \citep{makhzani2013k}, which uses the TopK activation function to enforce sparsity. This function selects the $k$ largest latent activations and sets the rest to zero. The function is defined as:

\begin{equation}
\begin{aligned}
    z &= \text{TopK}(l),
\end{aligned}
\end{equation}
where \( l \in \mathbb{R}^{n} \) is the latent feature representation. The decoder remains unchanged.

As shown in~\cite{gao2024scaling}, TopK activation offers several advantages: it directly enforces sparsity by controlling the number of active units, eliminating the need for \(\ell_1\) regularization; simplifies model comparison by avoiding hyperparameter tuning; improves sparsity-reconstruction trade-offs, especially in larger models; and enhances feature interpretability by zeroing out small activations.

\subsection{Diffusion Transformer }

Diffusion Transformers (DiT)\citep{dit} are transformer-based diffusion models operating in latent space, offering competitive image generation performance. Unlike traditional U-Net-based models, DiT leverages transformers to process latent patches, enhancing scalability and efficiency. In the latent diffusion framework (LDM)\citep{rombach2022high}, images are first encoded into latents via a variational autoencoder (VAE)~\citep{kingma2013auto}, and then a transformer models the denoising process, reducing computation while maintaining quality.

The forward diffusion process follows a Markov chain, sequentially adding Gaussian noise. At timestep $t$, the noisy data $x_t$ is formulated as:
\begin{equation}
x_t = \sqrt{\bar{\alpha}_t}x_0 + \sqrt{1-\bar{\alpha}_t}\epsilon_t, \quad \epsilon_t \sim \mathcal{N}(0, \mathbf{I}),
\end{equation}
where $\bar{\alpha}_t$ defines the noise schedule. The reverse process, modeled by the transformer, predicts noise $\epsilon_\theta(x_t)$ to iteratively denoise $x_t$. The training objective minimizes the mean squared error between predicted and true noise:
\begin{equation}
L_{\text{simple}}(\theta) = \mathbb{E}_{x_0, \epsilon_t, t} \left[\|\epsilon_\theta(x_t) - \epsilon_t\|^2\right].
\end{equation}

DiT's scalable architecture and strong performance on benchmarks like ImageNet position it as a advanced model for high-resolution image generation.

\section{Methodology}
\label{Sec:Method}

\subsection{Where to Train Sparse Autoencoders (SAEs)}  

Recent research on the mechanistic interpretability of diffusion models~\citep{basu2023localizing, basu2024mechanistic} has revealed that distinct cross-attentaion blocks are responsible for generating particular visual elements, including style and objects. Using these insights, since diffusion models share similar architectures, we selected a representative model, \textbf{PixArt-XL-2-1024-MS}~\cite{chen2023pixart} 
, to extract activation layers for training SAEs. A dedicated SAE was trained for each activation layer. The training process is in Fig.~\ref{fig:train_progress}

\subsection{Activations in DiTs}
To enable the use of latent activations from diffusion models for training SAEs, activations are extracted from specific transformer layers during both the forward diffusion and reverse denoising processes.

Given an input image $\mathbf{x}_0$ and a text prompt $\mathbf{p}$, the image is encoded into the latent space $\mathbf{z}_0$ using a VAE encoder. Noise $\boldsymbol{\epsilon}$ is added to simulate the diffusion process, producing a noisy latent $\mathbf{z}_t$ at timestep $t$. Simultaneously, the text prompt is embedded into $\mathbf{e}_p$ via a text encoder. The noisy latent $\mathbf{z}_t$ and text embedding $\mathbf{e}_p$ are fed into the transformer model, where hooks capture activations from specified layers during the forward pass. These activations are then used to train SAEs.

This approach leverages intermediate representations from the generation pipeline for downstream tasks. The activation extraction process is detailed in algorithm below:


\begin{algorithm}
\caption{Extract Activations from DiTs}
\begin{algorithmic}[1]
\Require Image $\mathbf{x}_0$, Text Prompt $\mathbf{p}$, Time step $t$, DiTs $\mathcal{T}$, Target layer $\ell$
\Ensure Activations $\mathbf{a}_\ell$
\State Encode image: $\mathbf{z}_0 \leftarrow f_{\text{enc}}(\mathbf{x}_0)$
\State Sample noise: $\boldsymbol{\epsilon} \sim \mathcal{N}(0, \mathbf{I})$
\State Add noise: $\mathbf{z}_t \leftarrow \sqrt{\bar{\alpha}_t}\mathbf{z}_0 + \sqrt{1 - \bar{\alpha}_t}\boldsymbol{\epsilon}$
\State Encode text: $\mathbf{e}_p \leftarrow f_{\text{text}}(\mathbf{p})$
\State Register hook at layer $\ell$: $g_\ell: \mathbf{a}_\ell = \phi_\ell(\mathbf{u}_\ell)$
\State Forward pass: $\mathbf{h}_{\text{out}} \leftarrow \mathcal{T}(\mathbf{z}_t, \mathbf{e}_p, t)$
\State Extract activations: $\mathbf{a}_\ell \leftarrow g_\ell(\mathbf{u}_\ell)$
\State \Return $\mathbf{a}_\ell$
\end{algorithmic}
\end{algorithm}

\vspace{-0.5cm}

\subsection{TIDE: Temporal-Aware SAE}
\label{subsec:tasae}

\begin{figure}[t]
\centering
\setlength{\belowcaptionskip}{-15pt}  
  \includegraphics[width=0.9\linewidth]{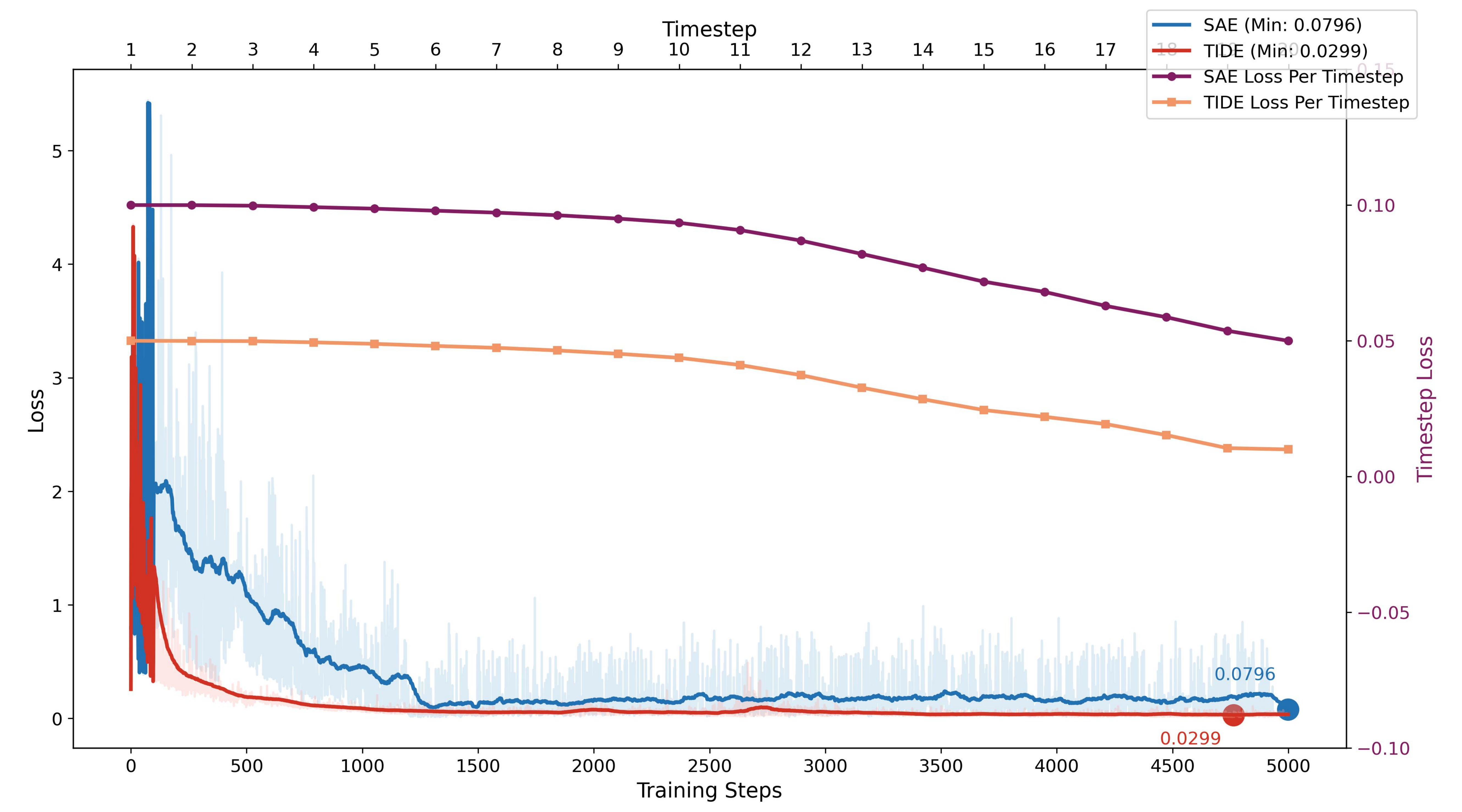}
  \caption{TIDE integrates timestep-dependent modulation into the original SAE architecture, achieving significantly faster convergence and enhanced performance.}
   \label{fig:ta-sae}
\end{figure}
In our training process, we observed that directly reconstructing the activation layers of DiTs using a SAE yielded relatively favorable results. However, the convergence was still noticeably slow. Inspired by the timestep sampling mechanism in diffusion models, we introduce temporal information into the activation layers to enhance the representation of timestep-dependent features.

The inclusion of temporal information is motivated by the fact that the token distribution varies significantly across different timesteps. For instance, at smaller timesteps (low $t$), the amount of information present is substantially different compared to larger timesteps (high $t$), where the input contains significant noise. To effectively disentangle these timestep-specific characteristics, it is crucial to inform the SAE of the corresponding timestep during reconstruction.

To address these timestep-dependent activation variations, we propose the Temporal-Aware SAE (TIDE), which incorporates a timestep-dependent modulation operation:
\begin{equation}
\begin{aligned}
x_\text{mod} = x \cdot (1 + \text{scale}(t)) + \text{shift}(t),
\end{aligned}
\end{equation}
where both scale(t) and shift(t) are functions of the current timestep $t$, modeled via a lightweight MLP initialized from the pre-trained adaptive Layer Normalization in DiT. This modulation allows the model to adaptively align activations across different diffusion timesteps.

As shown in Fig.~\ref{fig:ta-sae}, this modulation adjusts the activations by applying timestep-specific scaling and shifting, allowing the model to adapt to the varying distributions of activation layers across different timesteps.

Building upon the Top-k SAE framework, the addition of this modulation mechanism enables TIDE to effectively handle timestep-dependent activation distributions without requiring substantial modifications to the SAE parameters.

To prevent dead latents—latent variables that cease to activate during training—we employ careful initialization and revival strategies. Specifically, the encoder is initialized as the transpose of the decoder, and any inactive latents are periodically revived throughout the training process.

\section{Experiments}
\label{sec:exp}
\subsection{Training Setup}
\label{subsec:setup}

Each SAE/TIDE was carried out using three distinct strategies, with 40 hours of training on 8 A100 80G GPUs:  
\begin{itemize}
\item Direct Training: Using 80K prompt-image pairs extracted from CC3M~\cite{sharma2018conceptual} without augmentation.  
    \item Augmented Training: To mitigate overly dense local features and reduce overfitting, we randomly sample 1/16 of the tokens at the token level for each activation layer. Furthermore, we use a consistent set of 80K pairs to control for the effects of the sampling operation.
    \item TIDE Training: Incorporating a \textit{Temporal-Aware Architecture} to enhance the model's ability to capture temporal dynamics within the activation features.
\end{itemize}

\textbf{Evaluation}  
The performance of TIDE was evaluated using cosine similarity, mean squared error (MSE), and sparsity loss (see the supplementary material). These metrics also guided the training process, ensuring that the TIDEs(SAEs) struck a balance between sparsity and reconstruction accuracy.
A validation set of 500 prompt-image pairs is sampled from the CC3M dataset and is not used during training. For TIDE (SAE) models trained solely on the training data, this validation set is considered out-of-distribution (OOD).

\subsection{SAEs / TIDEs Exhibit Superior Reconstruction}
\label{subsec:reconstr}

\begin{table}[t]
\centering
    \caption{SAE ($d \times 16d$,sparsity 5\%, Avg of 28 layers) training Results (with/without Sampling Strategy and TA Architecture).}
  \begin{tabular}{@{}lcccc@{}}
    \toprule
    Train & $MSE$ & $S_{cos}$ & $MSE^{val}$ &$S_{cos}^{val}$\\
    \midrule
    SAE & $3.3e-3$&0.968&$1.23e-2$& \cellcolor{red!30}0.935 \\
    SAE(sample) & $2.5e-3$&0.970&$1.26e-2$& \cellcolor{green!30}0.962 \\
    TIDE & \cellcolor{blue!30}$3.1e-3$&0.970&$1.02e-2$& \cellcolor{red!30}0.944\\
    TIDE(sample) &\cellcolor{blue!30}$2.5e-3$& 0.972&$1.02e-2$& \cellcolor{green!30}0.964\\
    \bottomrule
  \end{tabular}
  \label{tab:train_sae}
\end{table}

We conducted experiments to evaluate the impact of token-level sampling during training and the integration of temporal-aware architecture into PixArt’s 28 Diffusion Transformer (DiT) activation layers. As shown in Tab.~\ref{tab:train_sae}, selecting only 1/16 of the total tokens during training significantly improves reconstruction accuracy on the training set (columns 1–2), while introducing only a minor increase in MSE. Notably, this strategy also leads to a marked improvement in cosine similarity on the out-of-distribution (OOD) validation set (column 4), suggesting enhanced generalization. In contrast, training without token sampling shows limited gains in reconstruction performance but fails to improve OOD alignment.

This result reveals two key insights:
\textbf{(1)} Token-level sampling improves learning efficiency by focusing on more informative local features, while also mitigating the impact of global noise—leading to better generalization.
\textbf{(2)} The negligible change in MSE indicates that the norm of the activation distribution remains stable, implying that the reconstruction process preserves essential structural information and does not adversely affect downstream image generation quality.

Furthermore, incorporating the TA architecture provides slight improvements in both MSE and cosine similarity, enabling more accurate reconstructions even at the $10^{-3}$ scale. This can be attributed not only to the sufficiently large latent space and carefully designed sparsity but also to the alignment with the principle of time embedding-based sampling in diffusion models.
\subsection{Scaling Laws and Downstream Loss of TIDEs}
\label{subsec:scaling}
\begin{figure*}[htbp]  
\setlength{\belowcaptionskip}{-15pt}  
  \centering
  \includegraphics[width=\textwidth]{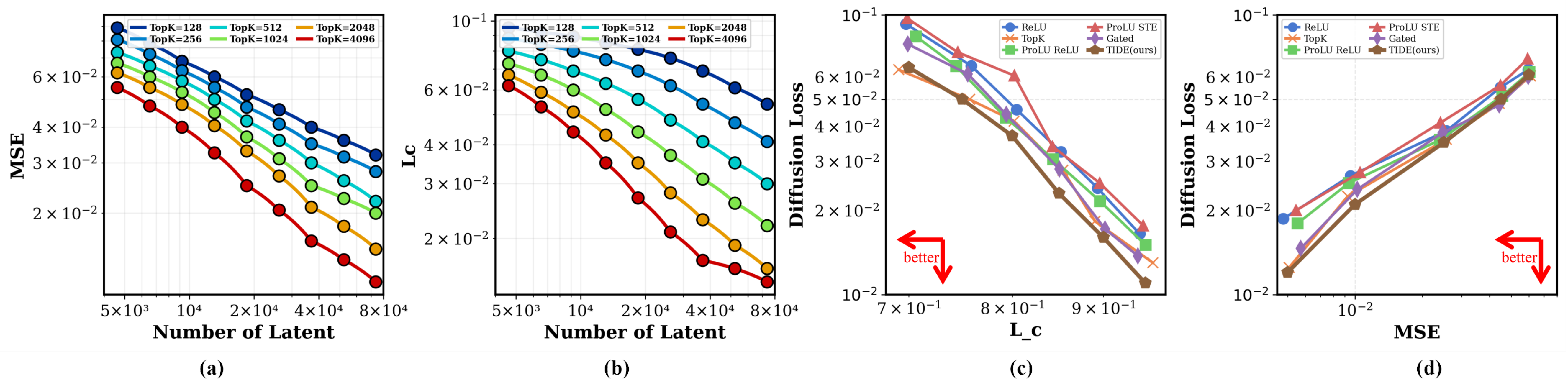}   
  \caption{
(a)(b) Scaling laws of convergence loss with fixed latents n under MSE and cosine similarity loss ($L_{cos} = 1 - S_{cos}$).
(c)(d) Comparison of TIDE and other activation functions: For 73728-d latents, TIDE achieves better trade-offs in diffusion loss against both cosine similarity and MSE.Based on these experiments, we selected top-k values in the range of 1024 to 4096 and set 16d = 73,728 as the parameters for subsequent image editing tasks.
}
  \label{fig:scaling}
\end{figure*}
Given the broad capabilities of advanced diffusion models — distinct from large language models (LLMs) — diffusion models exhibit significantly higher feature dimensionality and process a large number of tokens in each operation. Consequently, we hypothesize that accurately capturing the state of the model requires a large number of sparse features. To address this, we propose two primary approaches for determining the size of the autoencoder and the token budget:

As shown in Fig.~\ref{fig:scaling}(a)(b), increasing the number of latent dimensions from $d=4608$ to $16d=73728$, along with a higher top-k selection, leads to progressively better reconstruction of the activation layers. This improvement is reflected in a MSE reaching the order of $10^{-3}$ and a cosine similarity approaching 0.97. Compared to other diffusion U-Net-based SAEs, as well as most advanced LLM-based SAEs, the TIDEs trained using our strategy consistently achieve advanced performance.

We use the downstream diffusion loss to evaluate the reconstruction quality of TIDE. Specifically, the activation layers reconstructed by TIDE are used to replace the original activation layers in DiT, and the diffusion loss is then calculated. As shown in Fig.~\ref{fig:scaling}(c)(d),the diffusion loss of TIDE's reconstructed features is comparable to that of the original model and surpasses other types of SAEs, all while achieving advanced performance. This highlights TIDE's effectiveness in preserving features crucial to the generation task.
\subsection{Diffusion Really Learned Features}
\label{subsec: saereallylearn}

Recent studies have demonstrated the versatility of diffusion models across a range of downstream tasks, including depth estimation, classification, and segmentation. This raises two fundamental questions: 
\begin{enumerate}
    \item \textbf{Do diffusion models inherently encode the knowledge necessary for these tasks?}
    \item \textbf{Can TIDE validate the existence of such representations, thereby demonstrating their potential to support downstream applications?}
\end{enumerate}

To address these questions, we conducted a series of well-designed experiments to investigate the internal feature representations learned by diffusion models. For each target feature of interest, we selected 1K images containing identical target elements from a curated dataset optimized to minimize background interference and irrelevant semantic content. From these images, we extracted activations from the penultimate layer of the DiT model and applied average pooling to aggregate the latent features. Based on these pooled features, we computed similarity scores between the input image tokens and identified the top-k tokens with the highest similarity. The results, as illustrated in Fig.~\ref{fig:mask}, demonstrate that TIDE effectively exploits the internal representations of DiT to extract multi-level features, which can be categorized as follows:
\begin{figure}[h]  
    \setlength{\belowcaptionskip}{-15pt}  
  \centering
  \includegraphics[width=\linewidth]{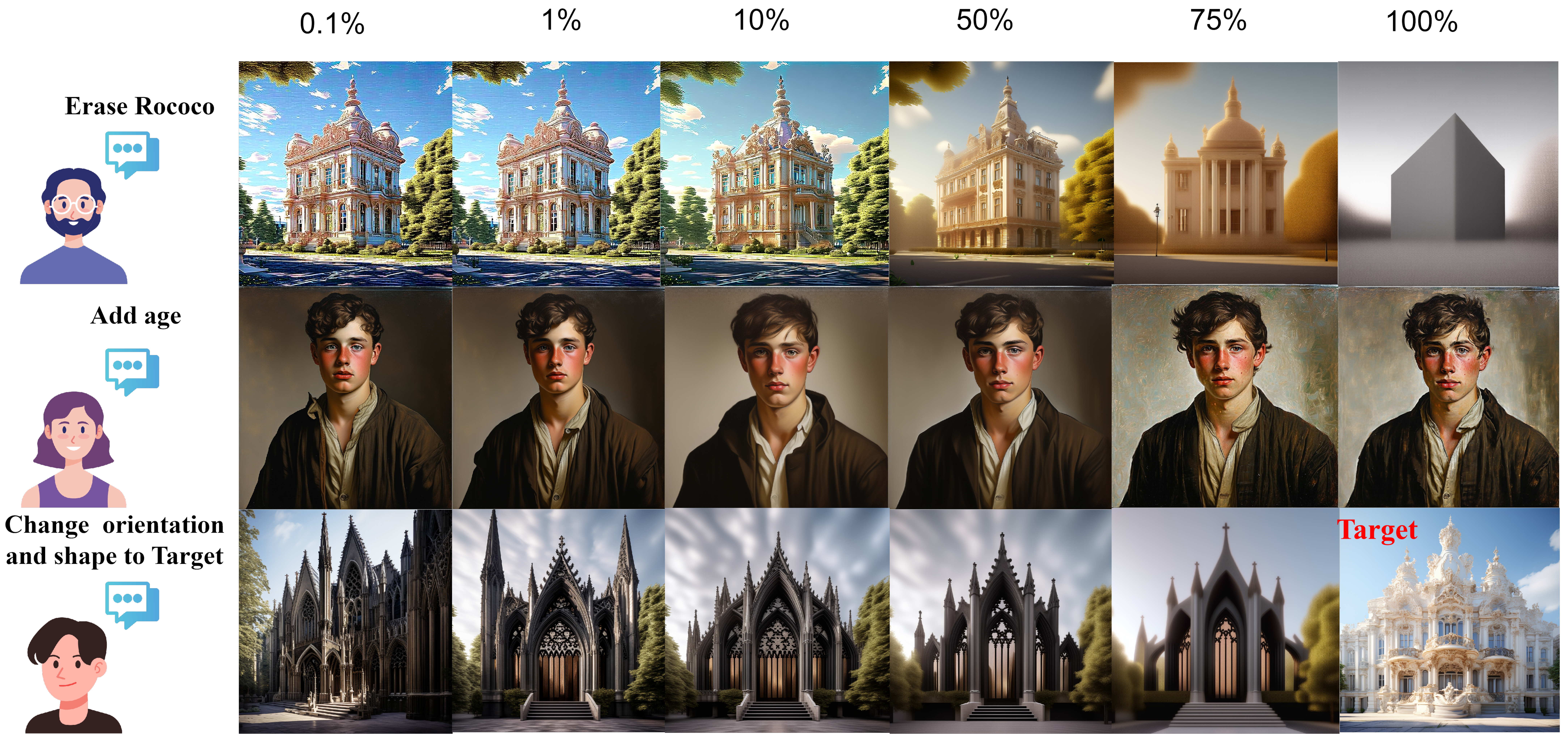}   
  \caption{By manipulating the latent space of the TIDE, we can achieve various concept transformations, such as erasing Rococo style, increasing age, and altering the orientation and shape of architecture. The extent of feature modification intensifies as the number of altered tokens increases. }
  \label{fig:application}
\end{figure}
\begin{itemize}
    \item \textbf{3D-Level}: TIDE captures features like “far,” “near,” and “shadow” that reflect foreground-background relationships. These features show highly specific activation patterns, and images with similar activations consistently share such 3D characteristics—highlighting the model’s ability to encode depth-related concepts useful for tasks like depth estimation.
    \item \textbf{Class-Level}: TIDE learns class-specific features and their subcomponents. For example, "Rococo" activates features like "house," "window," and "door," while "eagle" activates "bird," "fly," and "beak." These activations are localized, reflecting a clear hierarchical structure in the learned representations.
    \item \textbf{Semantic-Level}: At a finer granularity, TIDE identifies detailed semantic features and their spatial positions—e.g., pinpointing the "beak" of an eagle—demonstrating its capacity for extracting precise, localized semantic information.
\end{itemize}

As shown in Fig.\ref{fig:mask}, these results confirm that TIDE effectively encodes interpretable features across multiple levels—3D, class, and semantic—within pre-trained DiTs\cite{chen2023pixart}. This highlights the hierarchical structure of features learned during large-scale generative pre-training, demonstrating that diffusion models inherently capture and organize multi-level concepts. Such representations support diverse downstream tasks. For example, class- and semantic-level features enable classification~\cite{li2023your,mukhopadhyay2023diffusion} and segmentation, including zero-shot methods like DiffSeg~\cite{tian2024diffuse}. Likewise, the ability to learn 3D concepts allows these models to perform depth estimation, as shown in methods like Marigold~\cite{ke2024repurposing} and DepthFM~\cite{gui2024depthfm}.

\begin{table*}[h]
\centering
    \setlength{\belowcaptionskip}{-15pt}  
\caption{The use of TIDE achieves better editing performance without compromising the baseline quality. All results are evaluated on the MS-COCO dataset using FID (↓), ASR (↑) on I2P, and AlignScore (↑) under the T2I-CompBench benchmark.
The improvement in editing performance mainly stems from the noticeable increase in AlignScore, while the baseline quality remains stable as the FID only shows a marginal decrease.We highlight the best value in \colorbox{myblue}{blue} , and the second-best value in \colorbox{mygreen}{green}. }
\label{tab:ablation_combined}
\resizebox{\textwidth}{!}{
\begin{tabular}{ll|c|cc|ccc|cc|c}
\toprule
\multirow{3}{*}{\textbf{Setting}} & 
\multirow{3}{*}{\textbf{Configuration}} & 
\multirow{3}{*}{\textbf{ASR(\%) ↑}} & 
\multirow{3}{*}{\textbf{FID ↓}} & 
\multirow{3}{*}{\textbf{$\Delta$FID ↓}} & 
\multicolumn{6}{c}{\textbf{AlignScore ↑}} \\
\cmidrule(lr){6-11}
& & & & & 
\multicolumn{3}{c|}{\textbf{Attribute Binding}} & 
\multicolumn{2}{c|}{\textbf{Object Relationship}} & 
\multirow{2}{*}{\textbf{Complex}} \\
\cmidrule(lr){6-8} \cmidrule(lr){9-10}
& & & & & 
\textbf{Color} & \textbf{Shape} & \textbf{Texture} & 
\textbf{Spatial} & \textbf{Non-Spatial} & \\
\midrule
\multirow{7}{*}{\textbf{Hyper-parameters}} 
& \textbf{Baseline} (no SAE/TIDE)          & -- & 7.30 & --   & 0.688 & 0.549 & 0.708 & 0.210 & 0.318 & 0.412 \\
& SAE (5\% sparsity)              & -- & 7.93 & +0.63 & 0.602 &0.479&0.652&0.148&0.254&0.352 \\
& TIDE (5\%, 16d, default)        & -- & 7.45 & \cellcolor{mygreen}+0.15 & \cellcolor{mygreen}0.647 & \cellcolor{myblue}0.514&\cellcolor{mygreen}0.679&\cellcolor{mygreen}0.185&\cellcolor{myblue}0.271&\cellcolor{mygreen}0.392 \\
& TIDE (3\% sparsity)             & -- & 7.60 & +0.30 & 0.589 & 0.443&0.636&0.153&0.238&0.348 \\
& TIDE (latent dim = 4d)        & -- & 7.97 & +0.67 & 0.562 & 0.438&0.621&0.147&0.213&0.335 \\
& TIDE (latent dim = 8d)          & -- & 7.88 & +0.58 & 0.585 & 0.467&0.649&0.169&0.229&0.342 \\
& TIDE (latent dim = 32d)         & -- & 7.41 & \cellcolor{myblue}+0.11 & \cellcolor{myblue}0.672 & \cellcolor{mygreen}0.505&\cellcolor{myblue}0.696&\cellcolor{myblue}0.192&\cellcolor{mygreen}0.262&\cellcolor{myblue}0.407 \\
\midrule
\multirow{6}{*}{\textbf{DiT Layers}} 
& \textbf{Baseline} (no SAE/TIDE)          & -- & 7.30 & --   & 0.688 & 0.549 & 0.708 & 0.210 & 0.318 & 0.412 \\
& SAE (Block 27)                  & -- & 7.34 & \cellcolor{myblue}+0.04 & 0.663&\cellcolor{myblue}0.531&0.652&0.182&0.294&\cellcolor{myblue}0.382 \\
& TIDE (Block 4, early)           & -- & 10.49 & +3.19 & 0.495&0.337&0.475&0.102&0.152&0.141 \\
& TIDE (Block 10, mid)            & -- & 8.92 & +1.62 & 0.607&0.441&0.554&0.133&0.256&0.244 \\
& TIDE (Block 27, penultimate)    & -- & 7.64 & \cellcolor{mygreen}+0.34 & \cellcolor{myblue}0.672&\cellcolor{mygreen}0.537&\cellcolor{myblue}0.667&\cellcolor{myblue}0.195&\cellcolor{mygreen}0.301&\cellcolor{mygreen}0.352 \\
& TIDE (Block 28, final)          & -- & 7.34 & \cellcolor{myblue}+0.04 & \cellcolor{myblue}0.672&0.529&\cellcolor{mygreen}0.657&\cellcolor{mygreen}0.193&\cellcolor{myblue}0.309&0.349 \\
\midrule
\multirow{6}{*}{\textbf{Generalization}} 
& \textbf{SDXL}~\cite{podell2023sdxl}                 & -- & 8.24 & 0.00  & 0.638 & 0.544 & 0.528 & 0.212 & 0.318 & 0.417 \\
& SDXL + SAE                      & -- & 8.44 & +0.20 & \cellcolor{mygreen}0.583&\cellcolor{mygreen}0.504&\cellcolor{mygreen}0.451&\cellcolor{mygreen}0.167&\cellcolor{mygreen}0.295&\cellcolor{mygreen}0.348 \\
& SDXL + TIDE                     & -- & 8.38 & \cellcolor{mygreen}+0.14 & \cellcolor{myblue}0.607&\cellcolor{myblue}0.528&\cellcolor{myblue}0.495&\cellcolor{myblue}0.181&\cellcolor{myblue}0.309&\cellcolor{myblue}0.373 \\
& \textbf{FLUX-dev}~\cite{flux2024}                 & -- & 10.09 & 0.00  & 0.612 & 0.583 & 0.480 & 0.208 & 0.287 & 0.401 \\
& FLUX-dev + SAE                      & -- & 10.18 & +0.09 & \cellcolor{mygreen}0.489&\cellcolor{mygreen}0.441&\cellcolor{mygreen}0.432&\cellcolor{mygreen}0.178&\cellcolor{mygreen}0.216&\cellcolor{mygreen}0.314 \\
& FLUX-dev + TIDE                     & -- & 10.13 & \cellcolor{myblue}+0.04 & \cellcolor{myblue}0.549&\cellcolor{myblue}0.505&\cellcolor{myblue}0.453&\cellcolor{myblue}0.192&\cellcolor{myblue}0.247&\cellcolor{myblue}0.356 \\
\midrule
\multirow{6}{*}{\textbf{Safety}} 
& \textbf{SDv1.4}~\cite{Rombach_2022_CVPR}               & 17.80 & 16.40 & 0.00  & 0.310 & 0.336 & 0.347 & 0.344 & 0.350 & 0.339 \\
& ESD~\cite{Gandikota_2024_WACV}  & 2.84 & 18.08 & \cellcolor{mygreen}+1.68 & 0.315&0.34&0.351&0.347&0.355&0.343 \\
& SA~\cite{heng2024selective}                     & 2.78 & 24.96 & +8.56 & 0.319&0.344&\cellcolor{mygreen}0.355&\cellcolor{mygreen}0.354&0.359&\cellcolor{mygreen}0.348 \\
& RECE~\cite{gong2024reliable}                 & 0.80 & 17.02 & \cellcolor{myblue}+0.62  & \cellcolor{mygreen}0.321&0.337&0.348&0.345&0.351&0.34 \\
& SAFREE~\cite{yoon2024safree}                     & 1.48 & 19.88 & +3.48 & 0.316&0.341&0.342&0.348&0.356&0.344 \\
& CS~\cite{kim2025concept}                    & \cellcolor{myblue}0.42 & 18.59 & +2.19 & \cellcolor{myblue}0.327&\cellcolor{myblue}0.345&0.351&\cellcolor{myblue}0.368&\cellcolor{myblue}0.369&\cellcolor{myblue}0.357 \\
& TIDE (ours)              & \cellcolor{mygreen}0.64 & 19.72 & +3.32 & 0.317&\cellcolor{myblue}0.345&\cellcolor{myblue}0.357&0.352&\cellcolor{mygreen}0.366&0.33 \\
\bottomrule
\end{tabular}
}
\end{table*}

\vspace{-0.3cm}
\section{Potential Application}
\label{subsec:application}

Based on the analysis in the previous section, we choose to modify the activation layer of the penultimate DiT, which has a significant impact on the output. 
For each timestep, there always exists a directed vector between the latent feature and the target feature. At this point, the significance of Self-Attention Embedding is also reflected. By modifying the TIDE latent feature, we not only achieve better interpretability but also avoid out-of-distribution (OOD) errors caused by directly operating on the activation layer. Additionally, we can add the classifier-free guidance (CFG) Renormalization~\cite{lin2024stiv}, to ensure the stability of the norm.

The interpretable latent space enables various modification strategies, such as erasing or inverting top-K features to remove fine details, scaling activation indices to achieve feature transitions, or directly replacing latent features for global changes. These approaches can produce diverse effects, as shown in Fig.~\ref{fig:application}.


\vspace{-0.5cm}
\section{Ablation Study}
\label{sec:ablation}

To understand the effectiveness of our proposed \textbf{TIDE} module, we conduct comprehensive ablation experiments across four axes: 
\textbf{(1) hyper-parameter configurations, (2) DiT layer positioning, (3) generalization across diffusion backbones and (4) safe image editing}.

\vspace{0.5em}
\noindent\textbf{Evaluation Metrics.} We evaluate performance using three complementary metrics:
\begin{itemize}
    \item \textbf{Fréchet Inception Distance (FID)}~\cite{heusel2017gans}, using the FID-30K protocol on MS-COCO~\cite{lin2014microsoft}, to measure image generation quality.
    \item \textbf{Compositional Alignment}, via T2I-CompBench~\cite{huang2023t2i}, using AlignScore to evaluate semantic consistency with compositional prompts.
    \item \textbf{Safety}, using the I2P benchmark~\cite{schramowski2023safe}, reporting Attack Success Rate (ASR) to assess robustness against unsafe prompt injection.
\end{itemize}

\paragraph{Hyper-parameter Analysis}
We investigate the effect of key hyper-parameters in TIDE, including sparsity ratio, latent dimensionality, and token sampling.
As shown in Table~\ref{tab:ablation_combined}, although all TIDE variants exhibit slightly higher FID than the baseline, our default configuration (5\% sparsity, token sampling, 16d latent) leads to the smallest increase in FID (+0.15) and least drop in AlignScore across multiple dimensions, indicating minimal performance degradation.

Lowering the sparsity to 3\% further increases FID and reduces alignment, while disabling token sampling causes overfitting, reflected in a drop in both FID and semantic alignment. Increasing the latent dimension improves performance: 32d yields the highest Complex-score (0.407), though below the baseline. These results highlight a trade-off between sparsity and expressiveness, and suggest that moderate sparsity and sufficient latent capacity minimize performance impact.

\paragraph{Impact of DiT Layer Selection}
We examine the impact of injecting TIDE at different layers of the DiT backbone.
Table~\ref{tab:ablation_combined} shows that deeper layers yield better trade-offs: applying TIDE at the penultimate block (Block 27) results in the lowest FID increase (+0.34) and relatively strong alignment (e.g., 0.537 in Texture), making it the best-performing insertion point among all layers tested.

In contrast, injecting TIDE at earlier layers (e.g., 4) leads to significant performance degradation, with much higher FID (+3.19) and lower AlignScore. This confirms that deeper layers encode richer semantic features and are more suitable for sparse interpretability without harming image quality.

\paragraph{Generalization Across Diffusion Backbones}
To assess the model-agnostic nature of TIDE, we apply it to two additional diffusion backbones: SDXL and Flux-dev.
From Table~\ref{tab:ablation_combined}, we observe that while FID slightly increases in all variants, TIDE introduces the smallest changes: FID increases only from 8.24 to 8.38 on SDXL, and from 10.09 to 10.13 on Flux-dev. Meanwhile, alignment performance remains close to the baseline or shows modest improvement.

These findings suggest that TIDE generalizes well across architectures, maintaining stable generation quality with low overhead. While TIDE does not improve raw generation metrics, its primary benefit lies in adding interpretability with minimal performance penalty.

\paragraph{Safety Evaluation}
We further evaluate TIDE in the context of safe image editing using the I2P benchmark, which involves editing NSFW images to remove inappropriate content.
As shown in Table~\ref{tab:ablation_combined}, TIDE achieves a low Attack Success Rate (ASR) of 0.64\%, competitive with state-of-the-art defenses like CS (0.42\%) and SAFREE (1.48\%). Additionally, TIDE maintains generation quality (FID = 19.72), outperforming many baselines in safety–quality trade-offs.

Unlike the previous experiments, this setting involves actual image editing, where TIDE must not only preserve quality but also effectively sanitize unsafe content. Our results confirm that TIDE achieves strong safety guarantees with relatively minor degradation in visual fidelity.

These ablations demonstrate that:
\begin{itemize}
\item TIDE introduces minimal performance degradation, with the smallest increases in FID and smallest drops in AlignScore across hyper-parameters, layers, and backbones.
\item Moderate sparsity and 16–32d latent space offer the best trade-off between interpretability and performance.
\item Deeper DiT layers (especially Block 27) are the optimal points for interpretability injection.
\item TIDE generalizes well across diffusion backbones, confirming its robustness and portability.
\item In safe editing, TIDE filters NSFW content while preserving quality and defense.
\end{itemize}

\vspace{-0.2cm}

\vspace{-0.2cm}
\section{Conclusion}
\label{sec:conclusion}

TIDE introduces a sparse, temporally-aware interpretability framework for Diffusion Transformers, enabling transparent analysis of their internal representations. By uncovering multi-level semantic features, TIDE supports practical applications like image editing and demonstrates strong generalization across models and tasks. While current performance in complex editing scenarios remains limited by sparsity constraints, future work will explore more flexible sparse mechanisms to improve expressiveness. Overall, TIDE provides a solid foundation for building more trustworthy, controllable, and interpretable generative systems.

\bibliography{bibliography}

\end{document}